\documentclass[pmlr]{jmlr}

\usepackage{siunitx} 
\usepackage{comment}
\usepackage{longtable}
\usepackage{booktabs}
\usepackage{array}
\usepackage{multirow}
\usepackage{wrapfig}
\usepackage{float}
\usepackage{colortbl}
\usepackage{pdflscape}
\usepackage{tabu}
\usepackage{threeparttable}
\usepackage{threeparttablex}
\usepackage[normalem]{ulem}
\usepackage{makecell}
\usepackage{amsmath}
\usepackage{mathtools}
\usepackage{amsopn}
\usepackage{amsfonts}
\usepackage{caption}
\usepackage{pifont}
\usepackage{dsfont}
\usepackage{xcolor}
\usepackage{xspace}
\usepackage{hyperref}

\newcommand\ie{i.\,e.\xspace}

\newcommand\eg{e.\,g.\xspace}

\newcommand{\p}{\mathbb{P}} 
\newcommand{\E}{\mathbb{E}} 

\DeclareMathOperator*{\argmax}{arg\,max}

\makeatletter
\renewcommand{\fps@figure}{htb}         
\renewcommand{\fps@table}{htb}         
\makeatother

\usepackage{enumitem}
\setlist[enumerate]{leftmargin=.5in}
\setlist[itemize]{leftmargin=.5in}

\theorembodyfont{\upshape}
\theoremheaderfont{\scshape}
\theorempostheader{:}

\newtheorem{assumption}{Assumption} 

\makeatletter
\def\set@curr@file#1{\def\@curr@file{#1}} 
\makeatother

\theorembodyfont{\upshape}
\theoremheaderfont{\scshape}
\theorempostheader{:}
\theoremsep{\newline}

\title[Learning Optimal DTRs Using Causal Tree Methods]{Learning Optimal Dynamic Treatment Regimes\titlebreak Using Causal Tree Methods in Medicine}

\author{\Name{Theresa Bl{\"u}mlein}
       \Email{theresa@bluemlein.info} \\ 
       \addr ETH Zurich \\
       Zurich, Switzerland
       \AND
       \Name{Joel Persson}\thanks{Corresponding author}
       \Email{jpersson@ethz.ch} \\ 
       \addr ETH Zurich \\
       Zurich, Switzerland 
       \AND
       \Name{Stefan Feuerriegel}
       \Email{feuerriegel@lmu.de} \\ 
       \addr LMU Munich \\
       Munich, Germany
       } 
       
\begin{document}

\maketitle

\begin{abstract}
Dynamic treatment regimes (DTRs) are used in medicine to tailor sequential treatment decisions to patients by considering patient heterogeneity. Common methods for learning optimal DTRs, however, have shortcomings: they are typically based on outcome prediction and not treatment effect estimation, or they use linear models that are restrictive for patient data from modern electronic health records. To address these shortcomings, we develop two novel methods for learning optimal DTRs that effectively handle complex patient data. We call our methods \textsc{DTR-CT} and \textsc{DTR-CF}. Our methods are based on a data-driven estimation of heterogeneous treatment effects using causal tree methods, specifically causal trees and causal forests, that learn non-linear relationships, control for time-varying confounding, are doubly robust, and explainable. To the best of our knowledge, our paper is the first that adapts causal tree methods for learning optimal DTRs. We evaluate our proposed methods using synthetic data and then apply them to real-world data from intensive care units. Our methods outperform state-of-the-art baselines in terms of cumulative regret and percentage of optimal decisions by a considerable margin. Our work improves treatment recommendations from electronic health record and is thus of direct relevance for personalized medicine.
\end{abstract}

\section{Introduction}

Personalized medicine has emerged as a promising approach for personalizing treatment decisions to patients \citep{JMIR.2021, Kosorok.2019, Zhao.2013}. This is in contrast to the previous ``one-size-fits-all'' paradigm where patient heterogeneity in treatment responses and risk factors is largely ignored. Personalized medicine offers large benefits, in particular for chronic diseases as these oftentimes require multiple treatments over time \citep{Chakraborty.2013}. 

Prior literature has developed methods for learning sequences of individualized treatment decisions. One stream has emerged from statistics and is subsumed under \textbf{dynamic treatment regimes (DTRs)} \citep{Chakraborty.2013, Moodie.2007, Tsiatis.2019}. DTRs take historical patient information as input and output personalized treatment decisions. Classical DTR methods have relied on linear regression \citep{Chakraborty.2013, Murphy.2003, Murphy.2001, Tsiatis.2019}. These methods benefit from being explainable, yet, because of their simple structure, their use for modern electronic health records is limited. This is the case as such data are typically heterogeneous and embed unknown non-linear relationships between variables. This requires flexible and data-driven models. Another stream of works originates from machine learning, but these methods are primarily black-box (\eg, \citet{Liu.2017}) and, in general, are thus not suitable for personalized medicine. Hence, methods are needed for learning optimal DTRs that are explainable yet able to accommodate complex relationships.\footnote{In this paper, we refer to methods that reduce the opaqueness of a DTR as ``explainable''. We are aware that some authors distinguish ``interpretable'' (when the decision logic itself is transparent) vs. ``explainable'' (when insights are based on a surrogate model), see \citet{Amann.2020} and references therein for an overview. In this terminology, our \textsc{DTR-CT} would be classified as interpretable, while \textsc{DTR-CF} as explainable.}

In this paper, we develop novel methods, namely \textsc{DTR-CT} and \textsc{DTR-CF}, for learning optimal DTRs. Our proposed methods are based on a data-driven estimation of heterogeneous treatment effects using causal tree methods, specifically causal trees and causal forests. For this purpose, we adapt causal trees \citep{Athey.2016} and causal forests \citep{Athey.2018, Wager.2018} from the static setting to the sequential setting in DTRs. As a result, \textsc{DTR-CT} and \textsc{DTR-CF} help in making treatment recommendations from electronic health records. Importantly, our methods learn non-linear relationships, control for time-varying confounding, are doubly robust, and explainable.

We evaluate our methods as follows. (1)~We show the effectiveness using synthetic data. Here, our methods outperform baselines in terms of cumulative regret and decision accuracy (\ie, percentage of optimal decisions). Compared to the best performing baseline, \textsc{DTR-CT} and \textsc{DTR-CF} reduce the cumulative regret by more than 20\,\% in our synthetic data. In addition, \textsc{DTR-CT} and \textsc{DTR-CF} result in optimal treatment decisions at considerably higher rate. (2)~We demonstrate the applicability of our proposed methods to real-world medical data from intensive care units. We find that our methods recommend treatment decisions that result in better expected outcomes than under the observed treatment regime. 

Our main \textbf{contributions} are as follows\footnote{Our codes and results are publicly available via \url{https://github.com/jopersson/CausalTreeDTR}}:
\begin{enumerate}
    \item To the best of our knowledge, our work is the first to leverage causal trees and causal forests for learning optimal DTRs. Our proposed methods, \textsc{DTR-CT} and \textsc{DTR-CF}, thus learn heterogeneous treatment effects that are subject to non-linear relationships.
    \item Our proposed methods are explainable. For example, one can directly visualize the underlying decision tree in \textsc{DTR-CT}. Explainability is a key advantage for clinical relevance in medical decision-making and personalized medicine where the basis of treatment decisions to individual patients must be transparent and intelligible \citep{Amann.2020, Cutillo.2020, Ribeiro.2016}. 
    \item We outperform state-of-the-art baselines for learning optimal DTRs and demonstrate the applicability of our proposed methods to real-world medical data.
\end{enumerate}

\subsection*{Generalizable Insights about Machine Learning in the Context of Healthcare}

Our work also provides the following generalizable insights about machine learning for healthcare:
\begin{enumerate}
    \item Machine learning methods are increasingly developed that are both of high predictive accuracy and explainable \citep{Rudin.2019}. However, previous methods typically focus on \emph{predictions}, while we target \emph{decisions}. As such, we foresee important future work and impact in clinical practice through the use of prescriptive analytics for decision-making.
    \item We find that, for personalizing treatment decisions, classical statistical methods and prediction-focused machine learning methods may be sub-optimal. Instead, we see benefits from using machine learning methods that are explicitly developed for causal effect estimation.
    \item We empirically demonstrate that our proposed methods can improve on the treatment decisions of medical practitioners. This suggests that practitioners may benefit from adopting the methods in clinical practice. At the same time, this also points to untapped potential for further optimization of clinical decision-making in intensive care units.
\end{enumerate}

\section{Related Work}
\label{sec:related_work}

\textbf{Heterogeneous Treatment Effects.} The main goal of personalized medicine is to tailor treatments to patients by considering their heterogeneity. Doing so requires learning the treatment effects heterogeneity across patients. To quantify variation in treatment effects, methods to estimate heterogeneous treatment effects (HTEs) have recently gained in popularity. Examples include causal trees \citep{Athey.2016} and causal forests \citep{Athey.2018, Wager.2018}, which are designed for estimating HTEs in static settings but have not yet been adapted to learn optimal DTRs. 

\textbf{Dynamic Treatment Regimes.} Research on DTRs dates back to \citet{Robins.1997, Robins.2004}. Historically, these methods relied on linear low-dimensional parametric models estimated with least squares \citep{Moodie.2012, Murphy.2003, Murphy.2001, Tsiatis.2019, Wallace.2017}. Although the statistical properties of these methods are well-known, their simplicity may not be suited for modern electronic health record data that embed unknown non-linearities and complex relationships \citep{Liu.2017}.

\textbf{Machine Learning for DTRs.} Recently, machine learning methods have been proposed for accommodating the complexities in learning DTRs from electronic health record data. Although tree-based methods have been proposed for providing explainable treatment recommendations \citep{Speth.2021, Speth.2020, Sun.2021, Tao.2018, Zhang.2018, Zhao.2015}, they have been constructed as CART or random forests and thus do not explicitly model causal effects. However, the latter is a key property for reliably learning HTEs and, thus, to optimize treatment recommendations. Compared to previous tree-based methods, our proposed methods directly estimate HTEs with a causal interpretation via a tailored splitting criterion for the decision trees. As an alternative to DTRs, other works \citep{Cai.2021, Yauney.2018, Wang.2018, Prasad.2017, Raghu.2017} have used reinforcement learning. However, these methods are typically based on variants of deep learning or neural networks (see \citet{Yu.2021, Liu.2020} for surveys) and, therefore, are not readily explainable.

\textbf{Research Gap.} To the best of our knowledge, no previous study has used causal methods such as causal trees or causal forests to learn optimal DTRs. By leveraging these methods, our methods are able to better learn individualized treatment decisions from modern electronic health records data that are explainable to clinical practitioners.

\section{Setup}
\label{sec:setup}

\subsection{Causal Structure of the Data}

Let $i=1,\ldots,N$ refer to patients, and $t=1,\ldots,T$ refer to time steps. Each patient has $m$ baseline covariates $\mathbf C_i=[C_{i1},\ldots,C_{im}]$ (\eg, risk factors such as age or gender) measured prior to initial treatment, and has $p$ time-varying covariates $\mathbf X_{it}=[X_{it1},\ldots,X_{itp}]$ (e.g., vital sign measurements such as blood pressure or oxygen saturation) measured at every time step $t=1,\ldots,T$. Given the baseline covariates and time-varying covariates, a binary treatment $A^{(i)}_{t} = A^{(i)}_{t}(\mathbf{C}^{(i)}, \mathbf{X}^{(i)}_t) \in \{0,1\}$ is prescribed. Here, $A_{it} = 1$ means that patient $i$ is treated at time step $t$, and $A_{it} = 0$ that the patient is given no treatment or standard care. In the following, we will omit the patient superscript $(i)$ to simplify notation.

\begin{figure}
    \centering
    \includegraphics[width=0.4\linewidth]{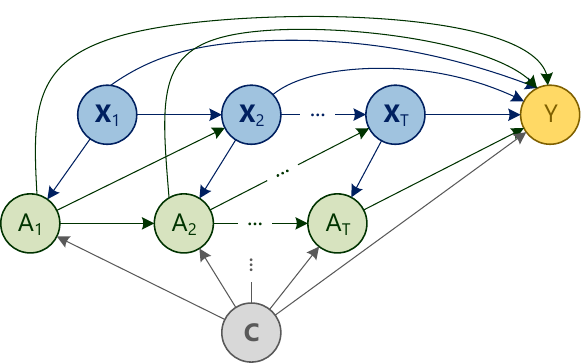}
    \caption{Causal structure of the sequential decision-making problem.}
    \label{fig:DGP}
\end{figure}

For brevity, let $A_{1:t}=(A_1,\ldots,A_t)$ denote the history of treatments up to and including time $t$, and let $\mathbf X_{1:t}=(\mathbf{X}_1,\ldots,\mathbf{X}_t)$ denote the time-varying covariates up to and including time $t$. Then, $\mathbf H_t \coloneqq (\mathbf{C}, A_{1:t-1}, \mathbf X_{1:t}) \in \mathcal H_T$ is the complete patient history through time $t$. The outcome of interest is $Y = Y(\mathbf H_T, A_T)$, which is observed at time step $T+1$ and depends on the full history (w.l.o.g. larger values are preferred). Fig.~\ref{fig:DGP} shows a directed acyclic graph of the the causal structure of data.

\subsection{Target Estimand: Optimal DTR}

Our estimands of interest are optimal DTRs, meaning rules for optimizing sequential treatment decisions. Let $\delta \coloneqq \left(\delta_1(\mathbf h_1),\ldots,\delta_T(\mathbf h_T)\right)\in\mathcal{D}$ denote an arbitrary DTR. Here, each $\delta_t \colon \mathcal H_t \to \{0,1\}$ is a mapping from the set of patient histories to the set of treatment alternatives at time $t$, and $\mathcal D$ is the set of all possible DTRs defined by a class of functions used to learn a DTR. Thus, a DTR $\delta$ is a sequence of decision rules that personalizes treatments to patients according to their heterogeneous histories.

Following standard practice in the causal inference literature, we use the potential outcomes framework \citep{Neyman.1923,Rubin.1974} to express our target estimand. Let $Y(\delta)$ denote the potential outcome that would be realized if treatments were assigned according to a DTR $\delta \in \mathcal D$. Moreover, let $V(\delta) \coloneqq \E[Y(\delta)]$ be the so-called \emph{value} of $\delta$, \ie, the mean outcome in the population if every patient was assigned treatments according to $\delta$. The target estimand is the optimal DTR $\delta^\text{opt}$, defined as 
\begin{equation}
    \delta^{\text{opt}} \coloneqq \argmax_{d\in\mathcal{D}} V(d).
\end{equation}

\subsection{Causal Inference for DTRs} 
\label{sec:ci}

To optimize sequential decisions, we must account for both prognostic and delayed effects of treatments as well as which decisions will be taken in the future. For learning optimal DTRs, \citet{Robins.2004} thus defined the so-called \emph{blip} function (from here on also called HTE) as the expected difference in potential outcomes at time $T$ would a patient be treated at time $t \leq T$ or not, conditional on the history and given subsequent optimal decisions. The blip function is thus given by
\begin{equation}
   \label{eq:ITE}
    \tau_t(\mathbf h_t) 
    \coloneqq 
    \mathbb E[Y(1, \delta^{\text{opt}}_{t+1:T}) - Y(0, \delta^{\text{opt}}_{t+1:T}) \mid \mathbf H_t=\mathbf h_t].
\end{equation} 
If the HTE was known, the optimal decision at time $t$ would be to assign treatment to the patients for whom treatment is beneficial, $\delta_t^{\text{opt}}=\mathds{1}\{\tau_t > 0\}$, assuming no treatment costs or restrictions. However, as we can only observe the factual outcome for the observed treatment regimen, the HTE is a counterfactual quantity. To identify it from data, we impose the following assumptions that are standard in the causal inference literature \citep{Robins.2008, Rosenbaum.1983}.

\begin{assumption}[Consistency] 
If $A_{1:T}=a_{1:T}$, then $Y(a_{1:T})=Y$.
\end{assumption}

\begin{assumption}[Stable Unit Treatment Value Assumption]
The potential outcome of any patient is independent of the treatment assignment of any other patient.
\end{assumption}

\begin{assumption}[Positivity]
If $\p(\mathbf H_t=\mathbf h_t)\not = 0$, then $\p(A_t=a_t \mid \mathbf H_t=\mathbf h_t) > 0$ $\forall$ $a_t\in\{0,1\}$, $t=1,\ldots,T$.
\end{assumption}

\begin{assumption}[Sequential Exchangeability]
$Y(a_t)\perp A_t \mid \mathbf H_t$ $\forall$ $a_t\in\{0,1\}$, $t=1,\ldots,T$.
\end{assumption}

\noindent
Assumption~1 links the potential outcomes to the observed outcomes by requiring that the two are equal under the same treatment assignments. Assumption~2 implies that there is no interference between treatment assignment and outcomes across patients. Assumption~3 states that, for any possible patient history, treatment assignment is not deterministic. Assumption~4 implies that conditioning on the patient histories is sufficient to remove confounding bias in estimated HTEs.

\section{Methods}

\subsection{Overview}
\label{sec:overview}

We now introduce our algorithms \textsc{DTR-CT} and \textsc{DTR-CF}. Therein, we combine a (i)~backward recursive estimation algorithm and (ii)~sequential HTE estimation. For the latter, we use a causal tree for \textsc{DTR-CT} and a causal forest for \textsc{DTR-CF}. In the following, we first introduce our backward recursive estimation algorithm (Section~\ref{sec:dtr-implementation}), which we then for our methods \textsc{DTR-CT} (Section~\ref{sec:dtr-ct}) and \textsc{DTR-CF} (Section~\ref{sec:dtr-cf}).

The rationale for using a backward recursive algorithm is the following. Backward recursion is a dynamic programming method used for learning optimal DTRs \citep{Murphy.2003} that addresses two fundamental challenges in sequential decision-making: (1) We must account for both prognostic and delayed effects of past decisions and the current decision \citep{Jiang.2022}. (2) Time-varying covariates can be both confounders and mediators of treatment effects depending on their time of measurement in relation to the treatment decision. Hence, conditioning on covariates measured after the treatment in a time step induces a post-treatment bias by ``explaining-away'' the treatment effect \cite[Ch.\,3.4.2]{Chakraborty.2013}. Thus, fitting a single model to the complete patient history leads to biased DTRs. 

Backward recursion solves the above-mentioned challenges by decomposing the sequential decision-problems into a set of single-step problems. Each single-step problem involves fitting a so-called outcome model to estimate the HTE and then solving for the optimal decision at that time step. Because the outcome is only observed at the final time step, we rely on optimizing so-called \emph{pseudo-outcomes} at each time step: the potential outcomes would the patient be treated optimally at later time steps \citep{Chakraborty.2013}.

\subsection{Backward Recursive Estimation Algorithm}
\label{sec:dtr-implementation}

\RestyleAlgo{ruled}

\begin{algorithm}[hbt]
 \hrulefill
 \caption{\textsc{DTR-CT} and \textsc{DTR-CF}}
 \label{alg:dtrtree}
 \vspace{0.25cm}
\footnotesize
\SetAlgoLined
\KwData{patient histories and outcomes $\{(\mathbf h_{T}, y)\}$} 
\KwResult{$\hat\delta^{\text{opt}} = (\hat\delta_1^{\text{opt}},\ldots,\hat\delta_T^{\text{opt}})\in\mathcal{D}$}

    Initialize DTR as the observed treatment sequence in the data: $\delta \leftarrow (a_1,\ldots,a_T)$.\;
    
    Initialize the pseudo-outcomes as the observed outcomes in the data: $Y'_T \leftarrow Y$.\;
    
    \For{$t=T,\ldots,1$}{
        Estimate propensity scores: $\hat{\pi}(\mathbf h_t) = (1+\exp(-\mathbf h^\top_t\hat{\gamma}_t)^{-1}$.
        
        \uIf{\textsc{DTR-CT}}{
            \begin{equation*}
                \hat{\tau}^{(\ell)}_t(\mathbf h_t)
                \leftarrow
            \textsc{DynamicWeightedCausalTree}(y \sim (\mathbf h_t, a_t), \text{weight} = (\hat{\pi}(\mathbf h_t))^{-1})
            \end{equation*}
            }
        \uElseIf{\textsc{DTR-CF}}{
            \begin{equation*}
                \hat{\tau}(\mathbf h_t)
                \leftarrow
            \textsc{DynamicWeightedCausalForest}(y \sim (\mathbf h_t, a_t), \text{weight} = (\hat{\pi}(\mathbf h_t))^{-1})
            \end{equation*} 
            }
         Estimate the optimal treatment decisions:
        \begin{equation*}
             \hat \delta^{\text{opt}}(\mathbf h_t) \leftarrow \mathds{1}\{\hat{\tau}(\mathbf h_t) > 0\}. 
         \end{equation*}
    
     Update the DTR: 
     \begin{equation*}
        \hat\delta \leftarrow (a_1,\ldots,a_{t-1},\hat\delta^{\text{opt}}(\mathbf h_t),\ldots,\hat\delta^{\text{opt}}(\mathbf h_T));
     \end{equation*}
     
     Update the pseudo-outcome via             
     \begin{equation*}
         Y'_{t-1} \leftarrow Y'_t + (\hat{\delta}^{\text{opt}}(\mathbf h_t) - a_t) \hat{\tau}(\mathbf h_t).
     \end{equation*} 
     } 
     Assign the estimated optimal decisions to the DTR: $\hat\delta^{\text{opt}} \leftarrow \hat\delta$\;
     
     \Return the estimated optimal DTR $\hat\delta^{\text{opt}}$\;

\hrulefill
\end{algorithm}

Algorithm~\ref{alg:dtrtree} presents our estimation procedure for our proposed methods. Algorithm~\ref{alg:dtrtree} takes patient histories and outcomes $(\mathbf h_1, \mathbf h_2, \ldots, \mathbf h_T, y)$ as input, and outputs an estimated optimal DTR from the set $\mathcal D$ of all possible DTRs defined by the chosen method (\ie, \textsc{DTR-CT} or \textsc{DTR-CF}). Depending on the method chosen, we fit a causal tree or causal forest, respectively, to estimate the HTE per time step. We do so by calling the functions \textsc{DynamicWeightedCausalTree} and \textsc{DynamicWeightedCausalForest}, which are causal trees and causal forests that we adapt from the static setting to the sequential setting for use in the backward recursive algorithm. Both are described in the subsequent sections.

Algorithm~\ref{alg:dtrtree} starts by initializing a DTR $\delta$ as the observed treatment sequence in the data, $\{a_1, a_2,\ldots, a_T\}$, and assigns the pseudo-outcome $y'$ as the observed final time step outcome $y$ for each individual. The recursion is then perform backwards, starting from the last time step and proceeding until the first, to update the DTR and pseudo-outcomes. 

For time step $t=T,\ldots,1$, the backward recursion proceed as follows:
\begin{enumerate}
    \item We estimate the so-called propensity scores \citep{Rosenbaum.1983} at time $t$. In our setting with binary treatments, the propensity score is defined as the conditional probability of treatment at time $t$ given the history, \ie, $\pi(\mathbf h_t) \coloneqq \p(A_t=1 \mid \mathbf H_t = \mathbf h_t)$. We estimate these using logistic regression of $A_t$ on $\mathbf H_t$ by solving $(1+\exp(-\mathbf h^\top_t\hat{\gamma}_t)^{-1}$ for maximum likelihood parameter estimates $\hat{\gamma}_t$. Nevertheless, in principle, any probabilistic classifier can be used.
    \item Depending on the chosen method (\ie, \textsc{DTR-CT} or \textsc{DTR-CF}), we estimate the HTE by fitting a causal tree or a causal forest to the observed outcomes $Y$ using the history $\mathbf h_t$. We follow the DWOLS method \citep{Wallace.2017} and weight the outcomes with the inverse of the estimated propensity scores. See Sections~\ref{sec:dtr-ct} (for \textsc{DTR-CT}) and \ref{sec:dtr-cf} (for \textsc{DTR-CF}) for details.
    \item We map the estimated HTE $\hat{\tau}(\mathbf h_t)$ to an optimal treatment decision $\hat \delta^{\text{opt}}(\mathbf h_t)$ by assigning the treatment if the HTE is positive, and assigning the control otherwise.
    \item By the plug-in principle, we update the DTR by imputing the estimated optimal decision at time $t$ for each individual.
    \item We update the pseudo-outcome for the next time step in the backward recursion. Here, the pseudo-outcome at the next iteration $t-1$ is updated as the pseudo-outcome at time $t$, plus the HTE at time $t$ if the observed and estimated optimal treatment decisions differ. Thus, the pseudo-outcome captures the increments (\ie, improvements) in the outcome under the estimated optimal treatment decisions. 
\end{enumerate}
Having completed the backward recursion, the algorithm outputs the estimated optimal DTR.

\subsection{DTR-CT}
\label{sec:dtr-ct}

We adapt causal trees \citep{Athey.2016} to the backward recursive algorithm for estimating optimal DTRs. Using a splitting criterion at every time step in the backward recursion, we create a partition of the patient population into subpopulations (\ie, a tree $\Pi_t$ consisting of leaves $\ell\in\mathcal L_t$, where $\mathcal L_t$ is a finite set of all leaves in time step $t$). The treatment effects are then estimated per heterogeneous leaf to obtain HTEs.

To ensure unbiasedness and avoid overfitting, we use the so-called \emph{honest} splitting criterion \citep{Athey.2016} for the partitioning that decouples model selection from model estimation. It does so by training the causal tree on a training set and estimating the HTEs with the trained causal trees on an estimation set.\footnote{We use the terms training set for model selection and estimation set for model estimation as in the original paper by \citet{Athey.2016}. In machine learning, the terms training set (for model estimation) and development set (for model selection) are also common.} The splitting criterion is then given by the expected mean squared error (EMSE) of the HTEs over the training set and the estimation set.

Let $\Pi_t = \{l_1,\ldots,l_{|\mathcal L_t|}\}$ denote a partitioning of histories $\{\mathbf h_t\}$ into $|\mathcal L_t|$ leaves such that $\bigcup_{\mathbf \Pi_t}\ell=\mathcal L_t$. Moreover, let $\mathcal S^{\text{tr}}_t$ and $\mathcal S^{\text{est}}_t$ denote the training set and estimation set, respectively, such that $\mathcal S^{\text{tr}}_t \cap \mathcal S^{\text{est}}_t = \emptyset$. Adapting the EMSE splitting criterion from the static setting to the sequential setting and for use with our backward recursive algorithm, we seek to find a partitioning that maximizes
\begin{equation}
\label{eq:EMSE-CT}
    -\text{EMSE}_{\tau_t(\mathbf h_t)}(\Pi_t)
    =
    \E_{\mathcal H_t} [\tau^2(\mathbf H_t; \Pi_t)]
    -
    \E_{\mathcal S^{\text{est}}_t,\mathcal H_t} [\mathbb V[\tau^2(\mathbf H_t; \mathcal S^{\text{est}}_t, \Pi_t)]].
\end{equation}
We further adapt its cross-validated sample estimator to our setting, given by
\begin{equation}
\label{eq:hatEMSE-CT}
    \begin{split}
        -\widehat{\text{EMSE}}_{\tau_t(\mathbf h_t)}(\mathcal S^{\text{tr,cv}}_t, N^{\text{est}}_t, \Pi_t)
        \coloneqq
        &\frac{1}{N^{\text{tr,cv}}_t} \sum_{i \in \mathcal S^{\text{tr,cv}}_t} \hat\tau^2(\mathcal H_t; \mathcal S^{\text{tr,cv}}_t, \Pi_t)
        -
        \left(
            \frac{1}{|\mathcal S^{\text{tr,cv}}_t|} + \frac{1}{|\mathcal S^{\text{est}}_t|}
        \right) 
        \cdot \\
        &\sum_{l \in \Pi_t}
        \left(
            \frac{S^2_{\mathcal S^{\text{tr,cv}}_{(A_t=1)}}(l)}{|\mathcal S^{\text{tr,cv}}_{(A_t=1)}|/N_t} + \frac{S^2_{\mathcal S^{\text{tr,cv}}_{(A_t=0)}}(l)}{1-|\mathcal S^{\text{tr,cv}}_{(A_t=0)}|/N_t}
        \right),
    \end{split}
\end{equation}
where $|\mathcal S^{\text{tr,cv}}_t|$ and $|\mathcal S^{\text{est}}_t|$ are the number of observations for model selection and model estimation, $|\mathcal S^{\text{tr,cv}}_{(A_t=1)}|$ and $|\mathcal S^{\text{tr,cv}}_{(A_t=0)}|$ the number of treated and control training observations, and $N_t$ the total number of observations until time $t$. Finally, $S^2_{\mathcal S^{\text{tr,cv}}_{(A_t=1)}}(l)$ and $S^2_{\mathcal S^{\text{tr,cv}}_{(A_t=0)}}(l)$ is the variance of the treated and control outcomes in leaf $l$. The superscript $(\text{cv})$ denotes the cross-validation fold. Intuitively, this splitting criterion rewards partitions with homogeneity within leaves and heterogeneity across leaves in terms of their treatment effects. See \citet{Athey.2016} for details. 

Having found a partitioning $\Pi_t$ that maximizes Eq.~\eqref{eq:hatEMSE-CT}, we estimate the HTE at time step $t$ in each leaf $\ell$ on the observed outcomes with
\begin{equation}
\label{eq:CT}
    \hat{\tau}^{(\ell)}_t(\mathbf h_t)
    =
    \frac{\sum_{\{i:A_{it}=0,\mathbf h_{it}\in \ell\}} \hat \pi^{-1}_{it} Y_i}{ \left|\{i:A_{it}=1,\mathbf h_{it}\in \ell\}\right|} 
    -
    \frac{\sum_{\{i:A_{it}=1,\mathbf h_{it}\in \ell\}}\hat \pi^{-1}_{it}Y_i}{\left|\{i:A_{it}=0,\mathbf h_{it}\in \ell\}\right|}.
\end{equation}
Here, $\mathbf h_{it}\in \ell$ denotes that patient $i$ with history $\mathbf h_{it}$ falls in leaf $\ell$ and $\hat \pi^{-1}_{it} = (\hat{\pi}(\mathbf h_{it}))^{-1}$ is the inverse of the estimated propensity score. In every leaf $\ell$, there are both treated and untreated patients. Hence, the time-specific HTE is estimated leaf-wise as the difference in average outcomes between treated and untreated patients \citep{Athey.2019}.

Algorithm~\ref{alg:causaltree} states the function \textsc{DynamicWeightedCausalTree} that we call inside the backward recursive algorithm. Both together yield our \textsc{DTR-CT}. 

\begin{algorithm}
\hrulefill
\caption{\textsc{DynamicWeightedCausalTree}}
\label{alg:causaltree}
\vspace{0.25cm}
\footnotesize
\KwData{patient histories, observed outcomes, and estimated propensity scores until time step $t$: ($\mathbf h_t, y, \hat \pi_t)$}
\KwResult{estimated HTE at time step $t$, \ie $\hat \tau_t^{(\ell)}$, for all $\ell \in \mathcal L_t$}

    Split the data into a training set $\mathcal S^{\text{tr}}_t$ and an estimation set $\mathcal S^{\text{est}}_t$ (for honesty)\;
    
    Find partitioning $\mathbf \Pi_t$ that maximizes Eq.~\eqref{eq:hatEMSE-CT} on $\mathcal S^{\text{tr}}_t$\;
   
    \Return partition $\mathbf \Pi_t$ of the training set into leaves $\mathcal L_t$\;
    
    \For{every leaf $\ell\in\mathcal L_t$}{
        Estimate the HTE at time step $t$, \ie, $\hat \tau^{(\ell)}_t(\mathbf h_t)$, on estimation set $\mathcal S^{\text{est}}_t$ using Eq.~\eqref{eq:CT}\;
    }
    \Return HTE at time step $t$, $\hat{\tau}^{(\ell)}(\mathbf h_t)$, for all $\ell \in \mathcal L_t$\;
    
\hrulefill
\end{algorithm}

\subsection{DTR-CF}
\label{sec:dtr-cf}

We now tailor causal forests to the sequential setting for with the backward recursive algorithm. A causal forest is an ensemble of causal trees \citep{Athey.2018,Wager.2018}, serving as an adaptive nearest-neighbor estimator by combining estimates from many causal trees. However, the HTE for a given patient is not estimated as a simple average of HTEs but as weighted average over the local patient neighborhood. Adapting the causal forests HTE estimator from static settings to use in our backward recursive algorithm, we get
\begin{equation}
\label{eq:CF}
    \hat{\tau}(\mathbf h_{it})
    =
    \frac{
        \sum_{j=1}^n \alpha_j(\mathbf h_{jt})
        (Y_j-\hat Y^{(-j)}) (A_{jt} - \hat \pi^{(-j)}_t)
        }
    { \sum_{j=1}^n \alpha_j(\mathbf h_{jt})(A_{jt}-\hat \pi^{(-j)}_t)^2}.
\end{equation} 
Here, $Y_j$ is the observed outcome for patient $j$, and $\hat Y^{(-j)} \coloneqq \hat \E[Y^{(-j)}\mid \mathbf H_{jt}=\mathbf h_{jt}]$ is its predicted value. We follow \citet{Athey.2018} and \citet{Athey.2019} and incorporate the propensity score estimate $\hat \pi^{(-j)}_t$ via the differences $(A_{jt}-\hat \pi^{(-j)}_t)$. Essentially, this corrects for lack of external validity due to unequal treatment assignment probabilities among patients. We also apply the honesty principle of predicting the outcome and propensity score ``out-of-bag'' not using data of patient $j$; hence, the subscript $(-j)$. Moreover,
\begin{equation}
    \label{eq:alpha}
    \alpha_j(\mathbf h_{jt})
    =
    \frac{1}{B}\sum_{b=1}^{B}\frac{\mathds{1}\{\mathbf h_{jt}\in\ell^{(b)}(\mathbf h_{jt})\}}{\mid\ell^{(b)}(\mathbf h_{it})\mid}
\end{equation}
is a data-adaptive kernel weight that reflects closeness in terms of covariate values, \ie, the frequency a patient $j\neq i$ falls into the same leaf $\ell^{(b)}$ as the patient of interest $i$. To construct the neighborhood for each single patient $i$, we calculate weights for every other patient $j\neq i$. See \citet{Athey.2019} for details on Eq.~\eqref{eq:CF} in the static setting.

Algorithm~\ref{alg:causalforest} states the function \textsc{DynamicWeightedCausalForest} that we call inside the backward recursive algorithm. Both together yield our \textsc{DTR-CF}.

\begin{algorithm}
\hrulefill
\caption{\textsc{DynamicWeightedCausalForest}}
\label{alg:causalforest}
\vspace{0.25cm}
\footnotesize
\KwData{patient histories, observed outcomes, and estimated propensity scores until time step $t$: ($\mathbf h_t, y, \hat \pi_t)$}
\KwResult{estimated HTEs at time step $t$, \ie, $\hat{\tau}(\mathbf h_t)$}
    
    \For{every tree $b=1,\ldots,B$}{
        Split the data into a training set $\mathcal S^{\text{tr},b}_t$ and an estimation set $\mathcal S^{\text{est},b}_t$ (for honesty)\;
        
        Find partitioning $\mathbf \Pi^{(b)}_t$ that maximizes Eq.~\eqref{eq:hatEMSE-CT} on $\mathcal S^{\text{tr},b}_t$\;
        
    \Return partition $\mathbf \Pi^{(b)}_t$ of the training set into leaves $\mathcal L^{(b)}_t$\;
        
    \For{every leaf $\ell\in\mathcal L^{(b)}_t$}{
        Estimate HTE at time step $t$, \ie, $\hat \tau^{(\ell,b)}_t(\mathbf h_t)$, on estimation set $\mathcal S^{\text{est,b}}_t$ using Eq.~\eqref{eq:CT}\;
    }
   \Return HTE at time step $t$, $\hat{\tau}^{(\ell)}(\mathbf h_t)$, for all $\ell \in \mathcal L_t$\;
    }
    Calculate weights $\alpha_i(\mathbf h_t)$ using Eq.~\eqref{eq:alpha}\;
    
    Calculate HTE at time step $t$, \ie, $\hat \tau_t(\mathbf h_t)$, as weighted average of $\{\hat \tau^{(\ell,b)}_t(\mathbf h_t)\}^B_{b = 1}$ using Eq.~\eqref{eq:CF}\;
    
    \Return HTE at time step $t$, \ie, $\hat \tau_t(\mathbf h_t)$\;

\hrulefill
\end{algorithm}

\subsection{Double Robustness}
\label{sec:doublerobust}

In our methods, we weight the observed outcomes with the inverse propensity score estimates when estimating the HTEs using causal trees or causal forests. This approach is analogous to the DWOLS method \citep{Wallace.2017} for learning optimal DTRs. With DWOLS, the HTEs are modeled as linear parametric regression functions estimated with weighted ordinary least squares, where the weights are given by the inverse of the estimated propensity scores. \citet{Wallace.2017} shows that this weighting procedure results in double robustness. Here, double robustness refers to that, per time step, only one of the propensity score model or HTE model needs to be correctly specified with unbiased estimators to get unbiased estimates of the HTE, and thus, the optimal decisions. In contrast to DWOLS, our proposed methods estimate the HTEs using causal trees and causal forests. As both of these unbiased estimators of HTEs \citep{Athey.2018, Wager.2018, Athey.2016}, and maximum likelihood is an unbiased estimator of logistic regression parameters, it thus follows that our proposed methods are also doubly robust.

\subsection{Explainability}

Both \textsc{DTR-CT} and \textsc{DTR-CF} are explainable:
\begin{itemize}
\item \textsc{DTR-CT}: At every time step, we can visualize the decision tree of which variables explain variation in the HTE across patients, and can thus be inspected by medical practitioners when deciding on treatments.
\item \textsc{DTR-CF}: At every time step, one can inspect the variable importance in the outcome model estimated with the causal forest, and thus, which variables are most prescriptive for treatment decision-making. 
\end{itemize}

\section{Simulation Studies}
\label{sec:simulation}

We perform simulation studies using synthetic data to demonstrate the effectiveness of our methods. Specifically, we compare the estimated optimal DTRs against a ground-truth (\ie, oracle) optimal DTR. 

\subsection{Data Generating Processes} 

We generate synthetic data analogous to previous works on DTRs \citep{Moodie.2012, Wallace.2015, Wallace.2017}. Specifically, we choose a similar number of patients histories, similar lengths for the patient histories, similar covariates. To this end, we simulate $N=5,000$ patient histories of length $T=3$. Each patient history consists of two baseline covariates $C_k\sim\mathcal{N}(\mu_k,1),\  k=1,2$, five continuous time-varying covariates generated as $X_{1,j}\sim\mathcal{N}(\mu_j,1)$ at $t=1$ and evolving as AR(1) processes $X_{t+1,j}\sim\mathcal{N}(X_{t,j} + \beta_j A_{t} ,2)$ over $t=2,3$ for $j=1,\ldots,5$, and two categorical time-varying covariates generated as $X_{1,j}\sim \mathrm{Binom}(N,3,p)$ at $t=1$ and evolving as AR(1) processes $\mathbf{X}_{t+1,j}\sim \mathbf{X}_{t,j} + A_t\cdot \mathrm{Binom}(N,3,p)$ for $t=2,3$ with $p_j$ being the success probability for $j=6,7$. The categorical covariates have four levels in the first time step $t=1$ and may have more than three levels at steps $1< t \leq 3$ due to the additive structure over the time steps. 

\textbf{Scenarios.} We consider two scenarios with varying complexity: (1)~linear interactions and (2)~non-linear interactions. Let $\tau_t^{(1)} \coloneqq f_t(\mathbf H_t)$ and $\tau_t^{(2)} \coloneqq f_t(g_t(\mathbf H_t))$ for $t=1,\ldots,T$ denote the true treatment effect under scenario 1 and 2, respectively. Here, $f_t(\cdot)$ is a function of arbitrary interactions between (predominantly categorical) patient history covariates, and $g_t(\cdot)$ is a non-linear function of the patient history. Hence, in scenario~1, the true treatment effect is a linear function of patient history interactions, whereas, in scenario~2, it is a non-linear function of their interactions. We construct the functions with a maximum interaction depth of three in both scenarios. We generate the true treatment effects such that the portion of optimally treated patients differs from the portion of actually treated patients. This mimics real-world settings where some patients are sub-optimally treated and, thereby, implies a potential for improvement by an optimal DTR. 

The outcome is defined as $Y^{(k)}\coloneqq\sum_{t=1}^T \phi^{(k)}_t(\mathbf{C}, \mathbf{X}_t) + A_t \cdot \tau^{(k)}_t+\varepsilon$, where $\phi^{(k)}_t(\cdot)$ is the scenario $k=1,2$ treatment-free function that determines the outcome in absence of treatment at time $t$, $A_t \cdot \tau^{(k)}_t$ is the treatment effect increment per scenario $k$ for the treated, and $\varepsilon\sim\mathcal{N}(0,1)$ is an error term. We define the true optimal treatment decision for scenario $k$ as $\delta^{\text{opt}}_{tk}=\mathds{1}\{\tau^{(k)}_t > 0\}$. The counterfactual outcome obtained under the true optimal DTR is then given by $Y^{\text{opt}}_k=\sum_{t=1}^T \phi^{(k)}_t(\mathbf{C}, \mathbf{X}_t) + \delta^{\text{opt}}_{tk}\cdot\tau^{(k)}_t$. For details, see Appendix~\ref{supp:dgp} and \texttt{data\_generating\_process.R} at \url{https://github.com/jopersson/CausalTreeDTR}.

\subsection{Baselines}

We compare our proposed methods to five state-of-the-art baselines for learning optimal DTRs from observational data (see Appendix~\ref{supp:overview} for details).

\textbf{Q-learning} \citep{Moodie.2012} estimates the HTEs for the treatment decision rules by modeling the potential outcomes separately as linear functions using OLS, and then taking their difference \cite[Ch.\,3.4.1]{Chakraborty.2013}.

\textbf{DWOLS} \citep{Wallace.2015} extends Q-learning by weighted ordinary least squares with inverse propensity score weights \citep{Wallace.2015}. Thus, this baseline is doubly robust. 

\textbf{G-estimation} \citep{Robins.2004} is another doubly robust regression-based estimator of optimal DTRs. It directly models the HTE as the difference in expected potential outcomes using generalized estimating equations \cite[Ch.\,4.3]{Chakraborty.2013}. 

\textbf{CART} \citep{Breimann.1984} use a splitting criterion that aggregates homogeneous patients in terms of their outcomes. We fit separate regression trees for the treated and the control groups and then estimate the HTEs as the mean difference in predicted outcomes between the groups. This serves as machine learning baseline to \textsc{DTR-CT} as it is also decision tree but not explicitly developed for the purpose of estimating causal HTEs.

\textbf{$k$-NN regression} \citep{Beygelzimer.2013} estimates the HTEs by taking the difference between the $k$-nearest patient in the treated and control group in terms of their history neighbourhood. As non-adaptive neighborhood estimator, and following previous research on causal forests \citep{Athey.2018}, it serves as reasonable baseline to \textsc{DTR-CF} that uses adaptive neighborhood matching. 

The first three baselines are classical DTR estimators from statistics built on semi-parametric regression functions. In contrast, \textsc{DTR-CT} and \textsc{DTR-CF} are non-parametric and data-driven machine learning methods. The last two baselines thus serve as machine learning baselines. While these are inherently non-parametric and data-driven, they lack in explainability. Explainability is a key advantage for clinical relevance in medical decision-making and personalized medicine where the basis of treatment decisions to individual patients must be transparent and intelligible \citep{Amann.2020, Cutillo.2020, Ribeiro.2016}. We thus only compare against baselines that are \textbf{explainable}. Moreover, we include those used in previous research on estimating HTEs (\ie, k-NN regression as a baseline for causal forests \citet{Athey.2018})

\subsection{Implementation Details}

We run Algorithm~\ref{alg:dtrtree} with \textsc{DTR-CT} and \textsc{DTR-CF}. For the baselines, we also use Algorithm~\ref{alg:dtrtree} but with the baselines instead of \textsc{DynamicWeightedCausalTree} or \textsc{DynamicWeightedCausalForest} to estimate the HTEs per time step, so that these output sequential treatment recommendations. For each method, we estimate the HTEs using the entire patient history $\mathbf H_t$ up until the time step in the backward recursion. Thus, all methods have access to all covariates. For Q-learning, DWOLS, and G-estimation, the outcome model is separated into two components (see Eq.~\eqref{eq:q-function} in Appendix~\ref{supp:classical}) that are regressed on $(\mathbf H_t, a_t\mathbf H_t)$. We use the same sets of covariates to estimate both component by inputting the entire patient history $\mathbf H_t$ into each. This mimics real-world applications where one typically controls for all pre-treatment variables in an attempt to satisfy sequential exchangability (Assumption~4 in Sec.~\ref{sec:ci}). We also estimate the baselines based on CART and $k$-NN regression as well as our \textsc{DTR-CT} and \textsc{DTR-CF} methods using the entire patient history up to the respective time step. For each method that use the propensity score, we estimate it in the same way by fitting a logistic regression of $A_t$ on $\mathbf H_t$ with parameter estimates obtained via maximum likelihood. For additional details on how we implemented the machine learning methods, see Appendix~\ref{supp:implementation}.

\subsection{Performance Evaluation}

For the synthetic data, we know the the oracle (\ie, true) optimal treatment decisions for benchmarking. We use 75\,\% of the generated data for training and 25\,\% for testing. Based on the optimal DTRs estimated on the training data, we predict optimal treatment decisions on the test data. We repeat estimation, prediction, and evaluation over $100$ runs, each time simulating new data from the data generating process. We compute the cumulative regret and decision accuracy on the test set using the true optimal DTR. For both quantities, we report mean and standard deviation (SD).

\textbf{Cumulative Regret.} The regret at time $t$ of an estimated optimal DTR $\hat{\delta}^{\text{opt}}$, denoted by $\rho_t(\hat{\delta}^{\text{opt}})$, is defined as difference between the (potentially suboptimal) estimated and the true optimal outcome $Y$, expressed in terms of the true HTE, \ie,
\begin{equation}
    \rho_t(\hat{\delta}^{\text{opt}})
    =
    (\delta_t^{\text{opt}} - \hat{\delta}^{\text{opt}}_t) \cdot \tau_t \geq 0, 
    \quad t=1,\ldots, T .
\end{equation}
The regret at any time step $t$ is always non-negative: $\rho_t$ equals zero if the estimated optimal decisions equal the true optimal decisions, and it is positive if they do not. The regret is zero (and thus minimal) if the estimated optimal decisions equals the oracle optimal decisions. Intuitively, the regret thus captures the extent to which estimated incorrect decisions impair the outcome. We obtain the cumulative regret for a given patient by taking the sum of her regrets over all time steps. We then average the patient-specific cumulative regrets over all patients and simulation runs. Smaller values are better.

\textbf{Decision Accuracy.} The decision accuracy describes the proportion of estimated optimal treatment decisions that match the true optimal treatment decisions from the oracle:
\begin{equation}
    \mathrm{acc}_t(\hat{\delta}^{\text{opt}}_t)
    =
    \frac{\mathds{1}\{\hat{\delta}_t^{\text{opt}}=\delta_t^{\text{opt}}\}}{N}
    \in[0,1],
    \quad t =1,\ldots,T.
\end{equation}
We average the decision accuracy over all time steps, patients, and runs. Larger values are better.

\subsection{Results}

We evaluate the different methods for learning optimal DTRs with regard to the cumulative regret (Table~\ref{tab:regrets}) and decision accuracy (Table~\ref{tab:accuracies}). Both \textsc{DTR-CT} and \textsc{DTR-CF} outperform the best-performing baseline approach by a large margin. For scenario~1 (linear interactions), \textsc{DTR-CT} and \textsc{DTR-CF} reduce the cumulative regret by 23.0\,\% and 27.1\,\%, respectively. For scenario~2 (non-linear interaction), both reduce the cumulative regret by 25.1\,\% and 48.2\,\%, respectively. In addition, \textsc{DTR-CT} and \textsc{DTR-CF} produce the true optimal treatment decisions in 7.2\,\% and 14.4\,\% more cases in the scenario~2 (non-linear interactions). 

\begin{table}[htb]
\resizebox{\linewidth}{!}{%
    \renewcommand{\arraystretch}{1.2}
\begin{tabular}{lrrrr}
\toprule
\textbf{Method} & \multicolumn{2}{c}{\textbf{(1) Linear interactions}} & \multicolumn{2}{c}{\textbf{(2) Non-linear interactions}} \\
\cmidrule(lr){2-3}\cmidrule(lr){4-5}
\textbf{} & \multicolumn{1}{c}{\textbf{Train}} & \multicolumn{1}{c}{\textbf{Test}} & \multicolumn{1}{c}{\textbf{Train}} & \multicolumn{1}{c}{\textbf{Test}} \\ \midrule
Q-learning \citep{Moodie.2012}        & 23.746 (0.532) & 23.730 (0.739) & 41.066 (1.630) & 41.487 (3.805) \\
DWOLS \citep{Wallace.2015}        & 23.491 (0.509) & 23.566 (0.821) & 40.756 (1.659) & 40.705 (1.990) \\
G-estimation \citep{Robins.2004}        & 55.747 (2.238) & 55.807 (2.462) & 20.362 (1.976) & 20.434 (2.068)\\
CART \citep{Breimann.1984}    & 4.096 (0.876) & 4.188 (0.956) & 25.388 (6.331) & 28.338 (26.967)  \\
$k$-NN Regression \citep{Beygelzimer.2013}        & 29.510 (0.872) & 26.005 (1.149) & 24.414 (2.508) & 31.750 (26.032)\\
 \textsc{DTR-CT}  (proposed)   & 3.204 (1.107) & 3.223 (1.131) & 15.200 (8.307) & 15.309 (8.471)  \\
  \textsc{DTR-CF} (proposed) & \textbf{3.035} (0.168) & \textbf{3.052} (0.197) &  \textbf{10.622} (1.496) & \textbf{10.582} (1.817)                 \\ \bottomrule
\multicolumn{5}{l}{Reported: Mean (SD). Lower values are better. Best performing approach in {bold.}}  
\end{tabular}
}
\caption{Cumulative regret of baselines and proposed methods for train and test set. }
\label{tab:regrets}
\end{table}

\begin{table}[htb]
\resizebox{\linewidth}{!}{%
    \renewcommand{\arraystretch}{1.2}
\begin{tabular}{lcccc}
\toprule
\textbf{Method}                                                      & \multicolumn{2}{c}{\textbf{(1) Linear interactions}}                          & \multicolumn{2}{c}{\textbf{(2) Non-linear interactions}}                      \\
\cmidrule(lr){2-3}\cmidrule(lr){4-5}
\textbf{}                                                              & \multicolumn{1}{c}{\textbf{Train}} & \multicolumn{1}{c}{\textbf{Test}} & \multicolumn{1}{c}{\textbf{Train}} & \multicolumn{1}{c}{\textbf{Test}} \\ \midrule
Q-Learning \citep{Moodie.2012}        & 69.5 (0.4) & 69.5 (0.7) & 56.3 (0.4) & 55.9 (0.8) \\
DWOLS \citep{Wallace.2015}        & 69.8 (0.4) & 69.8 (0.7) & 56.2 (0.8) & 56.2 (1.0)    \\
G-Estimation \citep{Robins.2004}        & 34.9 (2.5) & 34.8 (2.7) & 56.1 (1.2) & 56.0 (1.4)     \\
CART \citep{Breimann.1984}    &   82.7 (2.1)  & 82.5 (2.1) & 59.0 (3.4) & 58.4 (3.4)               \\
$k$-NN Regression \citep{Beygelzimer.2013}         & 64.3 (1.0) & 69.3 (1.2) & 54.7 (1.0) &	49.5 (1.7)                       \\
 \textsc{DTR-CT} (proposed)   & 83.9 (2.6) & 83.9 (2.6) & 65.8 (5.3) & 65.6 (5.4)     \\
 \textsc{DTR-CF} (proposed) & \textbf{82.6} (1.5)              & \textbf{82.6} (1.6)	& \textbf{72.7} (0.8) & \textbf{72.8} (1.0) \\ \hline
\multicolumn{5}{l}{Reported: Mean (SD). Higher values are better. Best performing approach in {bold.}}  
\end{tabular}
}
\caption{Decision accuracy [\%] of baselines and proposed methods for train and test set. }
\label{tab:accuracies}
\end{table}

\textbf{Sensitivity analysis.} The performance gains of our methods are robust to changes in hyperparameters. Specifically, we vary the minimum number of observations of treated and untreated patients per leaf between values of 10 and 70. We find little change in cumulative regret and decision accuracy, and that \textsc{DTR-CF} outperforms \textsc{DTR-CT} overall.

\begin{figure}
    \centering
    \includegraphics[width=0.5\linewidth]{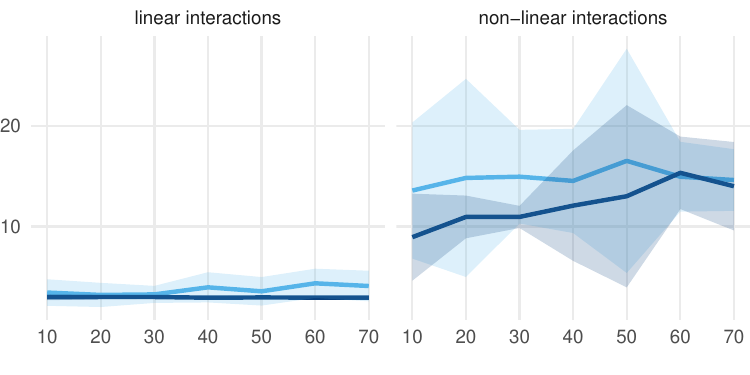}
    \includegraphics[width=0.5\linewidth]{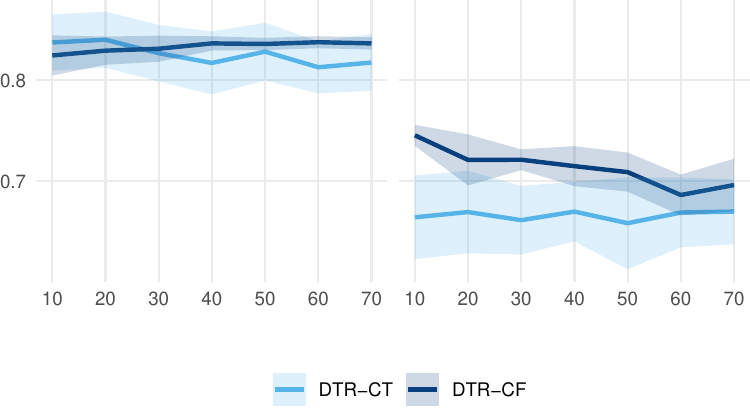}
    \caption{Cumulative regret (y-axis on top panels) and decision accuracy (y-axis on bottom panels) against the minimum number of treated and untreated patients per leaf (x-axis) of \textsc{DTR-CT} and \textsc{DTR-CF} in simulation studies with DGPs featuring linear interactions or non-linear interactions. The performance is evaluated on the test data over $100$ runs, each time with new data generated from the respective DGP. The point estimates are the mean across the runs. The shaded regions show the standard deviation across the runs.}
    \label{fig:tuning}
\end{figure}

\section{Empirical Application to Real-World Medical Data}
\label{sec:empirical_application}

\subsection{Setup}

We demonstrate the applicability of our proposed methods to real-world observational data of patients admitted to intensive care units (ICUs). Here, our aim is to prescribe mechanical ventilation or vasopressin over time such that length of stay is reduced. 

We use the MIMIC-III dataset and follow standard preprocessing \citep{Hatt.2021,Johnson.2016,Kuzmanovic.2021,Oezyurt.2021,Wang.2020}. We consider a smaller time of the stay in the ICU as proxy for patient health to be maximized. We control for seven time-varying covariates with vital signs: blood pressure [mean, diastolic, systolic], heart rate, oxygen saturation, respiratory rate, and temperature. We further control for six baseline covariates that may confound treatment effects: age, ethnicity, gender, race, insurance, admission type, and first care unit. As treatments, we consider mechanical ventilation and vasopressin. Both are key measures for life support, but also come with side-effects that oftentimes prolong the stay. 

We examine the patients who have been treated at least ten times, resulting in $N=8,059$ patients with histories over the first $T=10$ hours at the ICU. We imputed missing values of all time-varying covariates except temperature by their medians. We imputed missing values of temperature by a linear model using all other time-varying covariates as predictors as it exhibits the highest percentage of missing data amongst them. For \textsc{DTR-CT} and \textsc{DTR-CF}, we estimate the models using histories over three time steps to detect treatment effects delayed up to 3 time steps.
 
\subsection{Results}

For real-world observational data, the true optimal outcome and the true optimal decisions are never known. Hence, we cannot evaluate the performance of our methods against an oracle DTR, as we did in Sec.~\ref{sec:simulation}. Instead, we evaluate our methods by their ability to recommend treatment decisions that result in greater outcomes than those observed in the real-world medical data. The evaluation metric is given by
\begin{equation}
    r_t(\hat{\delta}_t^{\text{opt}}) = (\hat{\delta}_t^{\text{opt}} - a_t)\hat{\tau}_t , \quad t=1,\ldots,T.
\end{equation} 
The metric captures the expected gain in outcome to a given patient if they would be treated according to an estimated optimal DTR learned by one of our proposed methods instead of her factual treatment regimen. A larger value of the metric implies the treatment decisions of the estimated optimal DTR differs from the observed decision in the data and thus indicate an improvement.

\begin{table}[]
    \centering
    \resizebox{0.75\linewidth}{!}{%
       \begin{tabular}{lrr}
\toprule
\textbf{Approach}& \multicolumn{2}{c}{\hspace{1.2cm}\textbf{Treatment}}                          \\
& \multicolumn{1}{r}{\hspace{1.2cm}{Ventilation}} & \multicolumn{1}{r}{{Vasopressin}} \\ \midrule

\begin{tabular}[c]{@{}l@{}} \textsc{DTR-CT}  (proposed)\end{tabular}  & 2.176  & 2.097 \\
\begin{tabular}[c]{@{}l@{}}  \textsc{DTR-CF} (proposed)\end{tabular} & 0.233 & 0.203 \\ \bottomrule
\multicolumn{3}{l}{Cumulative expected rewards over time, averaged over patients.}  
\end{tabular}
    }
    \caption{Cumulative expected rewards for proposed methods on the MIMIC-III data.
    }
    \label{tab:mimic}
\end{table}

Overall, both \textsc{DTR-CT} and \textsc{DTR-CF} achieve a non-negative cumulative reward (see Table~\ref{tab:mimic}). Hence, both yield better estimated outcomes than the observed treatment regime. This suggests that our proposed methods offer improvements in patient care with real-world data. 

We refrain from comparisons to other DTR baselines as the estimated HTE is scale-variant and approach-specific. Rather, the purpose of the empirical application is to demonstrate the use and benefits of our methods on real-world data.

\subsection{Explainability}

Another advantage of our proposed methods is that they support standard tools for explainability.

\begin{figure}
    \centering
    \includegraphics[width=0.5\linewidth]{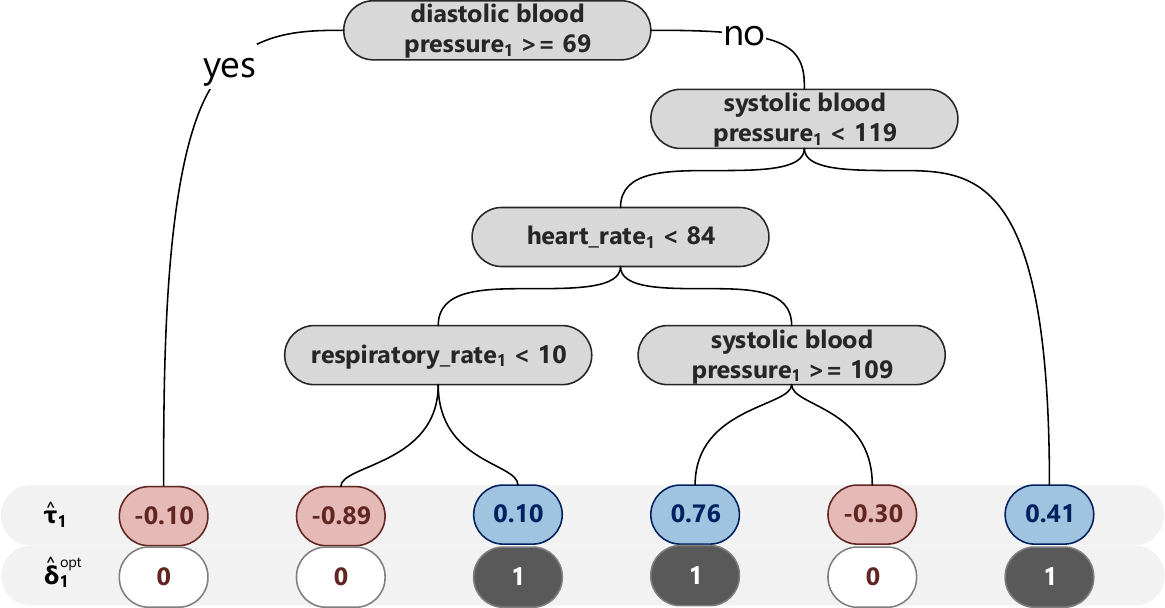}
    \caption{Estimated causal tree within \textsc{DTR-CT}.
    }
    \label{fig:causaltree}
\end{figure}

\textbf{Estimated Causal Tree.} Fig.~\ref{fig:causaltree} shows the estimated causal tree of \textsc{DTR-CT} of a specific time step. Patients within a leaf are homogeneous, but leaves are heterogeneous in terms of treatment effects. Every leaf shows the estimated treatment effect for its specific subgroup associated with the estimated optimal treatment decision $\hat{\delta}_1^{\text{opt}}$ and whether to apply ventilation. The variables responsible for splitting patients are the variables prescriptive of the recommended treatment decision in that time step. Hence, medical practitioners can compare the causal tree of an estimated optimal DTR against clinical practice and domain knowledge. Even if an estimated optimal DTR would not capture all nuances in clinical practice, it may still be useful for providing personalized decision support.

\begin{figure}
    \centering
    \includegraphics[width=0.4\linewidth]{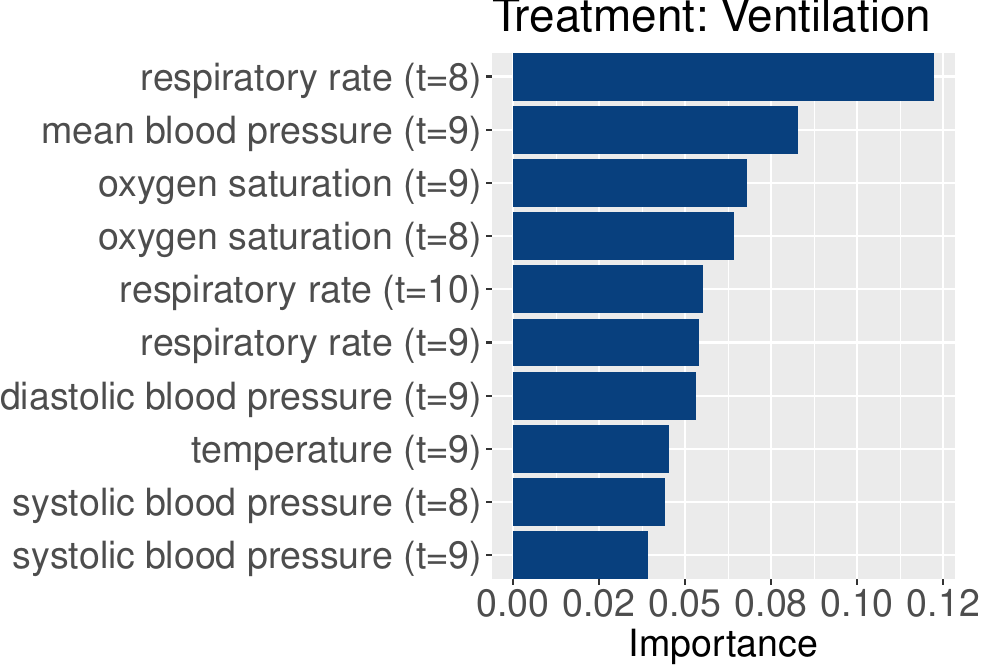}
    \includegraphics[width=0.405\linewidth]{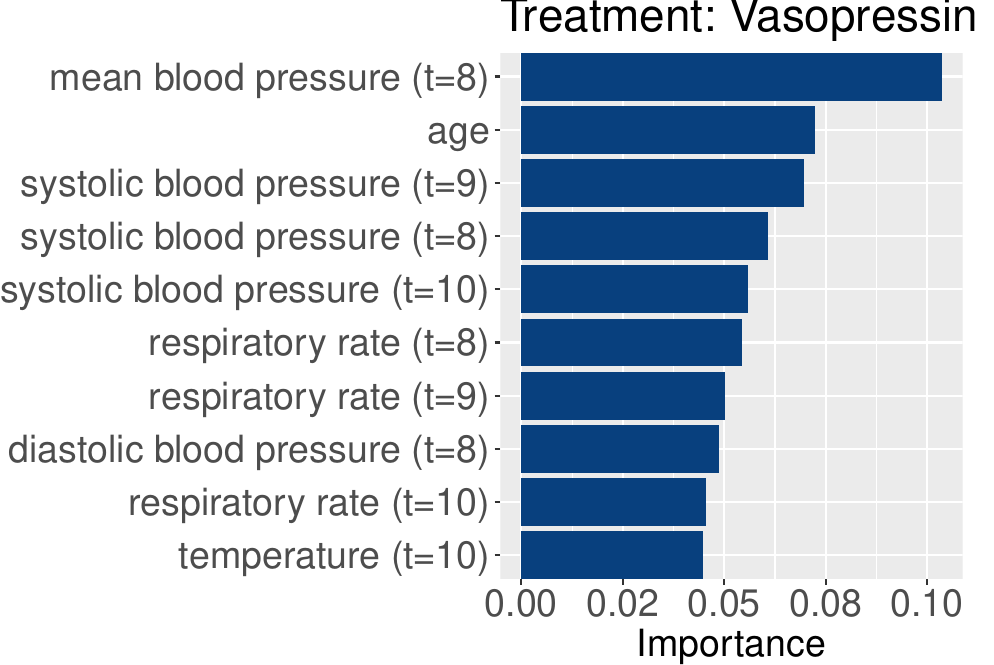}
    \caption{Variable importance within \textsc{DTR-CF} explaining which patient baseline covariates determine treatment assignment (here: $t = 10$).}
    \label{fig:variable_importance}
\end{figure}

\textbf{Variable Importance.} Fig.~\ref{fig:variable_importance} shows variable importance in the outcome model estimated with causal forest, and thus, for the estimated optimal treatment decisions by \textsc{DTR-CF}. We find that decisions to apply ventilation are largely based on patients' oxygen saturation, whereas to prescribe vasopressin, blood pressure is most important. This can be expected from medical knowledge and thus adds to the validity of our results.

\section{Discussion}
\label{sec:discussion}

In this paper, we propose two new methods for learning optimal DTRs for sequential treatment decision-making in medicine. Our methods build on causal trees and causal forests that we adapt from estimating static HTEs to estimating optimal DTRs in the sequential setting. The proposed methods flexibly learn non-linear relationships in patient data when personalizing treatment decisions, and account for additional patient heterogeneity by allowing the variables that are prescriptive of the optimal decisions to vary across patients. This is an advantage over classical baselines that assume the relationships to be known and the same variables to be prescriptive for all patients. 

Moreover, our methods achieve robust performance across different hyperparameter settings (see Fig.~\ref{fig:tuning}), implying little to no need for hyperparameter tuning. Finally, our methods address the need for explainability in medical practice. For instance, medical practitioners may inspect the estimated causal tree to understand which factors explain suggested treatment decisions and compare those against existing medical domain knowledge. This enables practitioners to better understand and validate optimal DTRs.

\paragraph{Limitations and Future Work} 

Our work is subject to limitations that open interesting paths for future work. First, we evaluate our methods over two time steps. This is consistent with prior works \citep{Moodie.2012, Wallace.2015, Wallace.2017}. Future work may investigate longer horizons, in case these are needed for a specific medical condition. Second, we explicitly benchmark our method against baselines that are also explainable. As such, our comparisons include linear models, tree models, and clustering. Explainability is a key requirement for clinical relevance in medical decision-making and personalized medicine where the basis of treatment decisions to individual patients must be transparent and intelligible \citep{Amann.2020, Cutillo.2020, Ribeiro.2016}. Third, we acknowledge that we demonstrated the effectiveness of our method through simulation studies. A promising path for future work is to implement our methods in the operational decision-making of healthcare providers to demonstrate the utility for patients in the field.  

\acks{We thank Milan Kuzmanovic for his introduction to MIMIC-III and his input to the empirical application. Joel Persson and Stefan Feuerriegel acknowledge funding from the Swiss National Science Foundation (grant 186932).}

\bibliography{references}

\clearpage

\appendix

\section{Details on Baselines for Learning Optimal DTRs}
\label{supp:overview}

Here, we describe the baseline methods against which we benchmark the perofrmance of \textsc{DTR-CT} and \textsc{DTR-CF}. Each of the baseline methods is embedded in Algorithm~\ref{alg:dtrtree} to estimate the HTE in place of causal trees or a causal forest. We first describe classical methods to learn optimal DTRs using methods originating from statistics, and then describe how we can estimate optimal DTRs using machine learning methods.

\subsection{Classical methods to Learning Optimal DTRs} 
\label{supp:classical}

Classical methods to learn optimal DTRs are, \eg, Q-learning \citep{Moodie.2012}, DWOLS \citep{Wallace.2015}, and G-estimation \citep{Robins.2004}. These typically makes use of linear regression-based models. At every time step $t=1,\ldots,T$, the potential outcome is modeled as the conditional expectation of the observed outcome given treatment and covariate history up until step $t$ and given optimal treatments thereafter \citep{Chakraborty.2013}, \ie,
\begin{equation}
    \label{eq:q-function}
    \E[Y \mid \mathbf H_t=\mathbf h_t,A_t=a_t]
    =\beta_t^\top \mathbf h_t + a_t(\psi_t^\top\mathbf h_t)\text{.}
\end{equation}
Here, $\beta_t^\top\mathbf H_t$ is the time $t$ treatment-free model that predicts the observed outcome in the absence of treatment in time $t$ and $A_t(\psi_t^\top\mathbf H_t)$ is a  treatment-interaction model, also known as a blip model, with which history covariates determine the heterogeneity in the effect of treatment at time $t$ on the observed outcome. 

The additive separation into a treatment-free model and a treatment-interaction model implies that treatment decisions are completely based on the latter. If the parameters $\beta_t$ and $\psi_t$ are known, the HTE at time $t$ is given by ${\tau}(\mathbf h_t)={\psi}_t^\top\mathbf h_t$. By the plug-in principle, its estimate is in turn $\hat{\tau}(\mathbf h_t)=\hat{\psi}_t^\top\mathbf h_t$. This gives an estimated optimal treatment decision 
\begin{equation}
    \label{eq:opt_decision}
    \hat \delta_t^{\text{opt}}(\mathbf h_t)=\mathds{1}\{\hat{\psi}_t^\top\mathbf h_t > 0\}.
\end{equation}
Note that although the parameters are estimated, these methods assume that the true model is low-dimensional, linear in parameters $\psi_t$, and has known functional form and variables. It also assumes that the same set of patient covariates are important for outcome prediction as for treatment effect heterogeneity and, thus, individualized treatment recommendations. In practice, these sets may be different and unknown.

\subsection{Q-Learning} 
\label{sec:qlearn} 
In Q-learning, the expected potential outcome is modeled as the conditional expectation of the outcome given treatment and covariate history \citep{Moodie.2012}. At every time step $t=T,\ldots,1$ in the backward recursion, this conditional expectation is separated into a treatment-free model and the treatment-interaction model according to Eq.~\eqref{eq:q-function}.

The blip parameters $\psi_t$ are estimated with ordinary least squares. Specifically, the closed-form solution is given by the least squares equation
\begin{equation}
    \hat{\psi}_t(\mathbf H_t,a_t)= \left((a_t\mathbf H_t)^{\top}(a_t\mathbf H_t)\right)^{-1}(a_t\mathbf H_t)^{\top} Y.
\end{equation} 
Given parameter estimates $\hat{\psi}_t = \hat{\psi}_t(\mathbf H_t,a_t)$ and using the plug-in principle, Q-Learning estimates the HTE at time step $t$ as $\hat{\tau}(\mathbf h_t)=\hat{\psi}_t^\top\mathbf h_t$, which is a linear function in the parameters $\hat{\psi}_t$ and patient history $\mathbf h_t$. For each patient at every time step, the estimated optimal treatment decision is then given by Eq.~\eqref{eq:opt_decision}.

\subsection{Dynamic Weighted Ordinary Least Squares} 
\label{sec:dwols}

Dynamic weighted ordinary least squares (DWOLS) extends Q-learning by weighting observations with the inverse propensity score estimate $\hat{\pi}^{-1}_t = (\hat{\pi}(\mathbf h_t))^{-1}$ \citep{Wallace.2015, Wallace.2017}. To estimate the blip parameter for time step $t=1,\ldots,T$, the observed outcome $Y$ is regressed on $(\mathbf H_t,a_t \mathbf H_t)$. The closed-form least squares solution is 
\begin{equation}
    \hat{\psi}_t(\mathbf H_t,a_t) 
    = 
    \left((a_t \mathbf H_t)^{\top}(\hat{\pi}^{-1}_t a_t\mathbf H_t)\right)^{-1}(a_t \mathbf H_t)^{\top} (\hat{\pi}^{-1}_t Y).
\end{equation}
The HTE is then estimated as $\hat{\tau}(\mathbf h_t)=\hat{\psi}_t^\top\mathbf h_t$, which, like in Q-learning, is a linear function in parameters and patient history. Also as for Q-learning, we use the same sets of history covariates to estimate the two models involved in DWOLS. The only difference to Q-learning is that by weighing observations with the inverse propensity score, DWOLS is doubly robust against model mis-specification.

\subsection{G-Estimation} 
\label{sec:gestimation}

Similar to Q-learning and DWOLS, G-estimation is a semi-parametric regression-based method to sequentially estimate HTEs using backward recursion \citep{Robins.2004}. However, instead of weighting observations with the inverse propensity score, like DWOLS, it directly incorporates the estimated propensity score $\hat{\pi}(\mathbf h_t)$ in the estimating equation, given by
\begin{equation}
\begin{split} 
    \hat{\psi}_t(\mathbf H_t,a_t) 
    = 
    \left(((a_t-\hat{\pi}_t) \mathbf H_t)^{\top}(a_t \mathbf H_t)\right)^{-1}
    \left((a_t-\hat{\pi}_t) \mathbf H_t\right)^{\top}Y.
\end{split} 
\end{equation} 
Using blip parameters obtained from the above estimating equation, the HTE estimate is given by $\hat{\tau}(\mathbf h_t)=\hat{\psi}_t^\top\mathbf h_t$.

\subsection{CART} 
\label{sec:cart}

CART \citep{Breimann.1984} (or simply, decision trees) use a splitting criterion that aggregates homogeneous patients in terms of their outcomes. We fit two separate regression trees, CART1 and CART0, for the conditional expectation of the observed outcome among the treated and the control groups, given the history and treatment, at every time step $t$. The estimated HTE is then given by their difference, \ie,
\begin{equation}
    \begin{split}
        \hat{\tau}^{(\text{CART})}_t(\mathbf h_t)
        =
        \hat{\mathbb E}_{\text{CART1}}\left[Y \mid \mathbf H_t=\mathbf h_t, A_t=1\right] 
        -
        \hat{\mathbb E}_{\text{CART0}}\left[Y \mid \mathbf H_t=\mathbf h_t, A_t=0\right].
    \end{split}
\end{equation}

\subsection{$k$-NN Regression} 
\label{sec:k-nn}

With $k$-NN regression, we estimate the HTE at the time step $t$ as the difference in conditional expectation of the observed outcomes for the treated and control groups, given the patient histories:
\begin{equation}
    \begin{split}
        \hat \tau^{({k\text{-NN}})}_t(\mathbf h_t)
        =
        \hat{\mathbb E}_{k\text{-NN1}}\left[Y \mid\mathbf H_t=\mathbf h_t, A_t=1\right] 
        -
        \hat{\mathbb E}_{k\text{-NN0}}\left[Y \mid\mathbf H_t=\mathbf h_t, A_t=0\right].
    \end{split}
\end{equation} 
Here, $k$-NN1 and $k$-NN0 denote $k$-nearest neighbor~(NN) regression models estimated on the treatment and control groups, respectively. The models are fitted on the $k$-nearest patients in each group within the neighbourhood of the patient of interest in terms of history covariate values $\mathbf h_t$. 

\section{Details on Data Generating Processes}
\label{supp:dgp}

\subsection{Scenario 1: Linear Interactions} 

The true treatment-free functions for $t=1,2,3$ are constructed via
\begin{align}
   \phi_1(\mathbf{X}_1) &\coloneqq X_{11} + X_{12} + X_{13} + X_{14}, \\
   \phi_2(\mathbf{X}_2) &\coloneqq X_{21} + X_{22} - X_{23} + X_{24}, \\
   \phi_3(\mathbf{X}_3) &\coloneqq X_{31} + X_{32} - X_{33} + X_{34}.
\end{align}

\subsection{Scenario 2: Non-Linear Interactions}

The true treatment-free functions for $t=1,2,3$ are constructed non-linearly via
\begin{align}
  \phi_t(\mathbf{X}_1) &\coloneqq 5 + X^2_{19} - 3 \sin (C^2_2 + X_{11} - X_{12} + X_{13}), \\
  \phi_t(\mathbf{X}_2) &\coloneqq \mu_1(\mathbf{X}_1) + X^2_{29} - 3 \sin (X_{21} - X_{22} + X_{23}), \\
  \phi_t(\mathbf{X}_3) &\coloneqq  \mu_2(\mathbf{X}_2) + X^2_{39} - 3 \sin (X_{31} - X_{32} + X_{33}).
\end{align}

See the main paper text and codes at our repository (\url{https://github.com/jopersson/CausalTreeDTR}) for details on how the data and HTEs are constructed.

\subsection{Details on Implementation of Machine Learning Baselines}
\label{supp:implementation}

\textbf{CART.} We set the minimum number of observations required in each leaf to $10$. This is half of the required number used in \textsc{DTR-CT} because we train two separate trees, whereas in \textsc{DTR-CT}, we train a single tree on treated and control patients at each time $t$. 

\textbf{$k$-NN Regression.} We set $k=10$ in each estimator, which corresponds to the minimum number of observations required in each causal tree in \textsc{DTR-CT}, \ie, 20. 

\textbf{DTR-CT.} We use 50\,\% of the patient population as training set and 50\,\% as estimation set at each step $t$. The minimum number of treated and control patients in each leaf of every causal tree is set to $20$ in total, consisting of $10$ patients from each subgroup. To prevent overfitting, the maximum number of buckets used for splitting is set to $20$. We run 5-fold cross-validation on Eq.~\eqref{eq:hatEMSE-CT} to optimize the pruning of the decision trees.

\textbf{DTR-CF.} We estimate each causal forest per time step in \textsc{DTR-CF} as an ensemble of 500 causal trees. We set the minimum number of treated and control patients in each leaf of every causal tree to $20$ in total, consisting of $10$ patients from each subgroup.

For details, see the codes at our repository (\url{https://github.com/jopersson/CausalTreeDTR}).

\end{document}